%% file: main.tex
\documentclass{article}
\PassOptionsToPackage{authoryear}{natbib}
\usepackage{crossfly,times}
\setcitestyle{authoryear,round}
\input{math_commands.tex}

\usepackage{amsmath,amssymb,graphicx,booktabs,multirow,array}
\usepackage{placeins}
\usepackage[table]{xcolor}
\usepackage{url}
\usepackage{enumitem}
\usepackage[T1]{fontenc}
\usepackage{float}
\usepackage{pifont}
\usepackage{xcolor}
\usepackage{wrapfig}
\usepackage{hyperref}

\definecolor{mydarkblue}{rgb}{0,0.08,0.45}
\hypersetup{
    colorlinks=true,
    citecolor=mydarkblue,
    linkcolor=mydarkblue,
    urlcolor=magenta
}

\newcommand{\cmark}{\ding{51}}%
\newcommand{\xmark}{\ding{55}}

\definecolor{mygray}{gray}{0.90}

\newcommand{\thickhline}{\noalign{\hrule height 1pt}}

\newcommand{\projectname}{\textsc{CrossFly}}

\title{\makebox{Benchmarking Interaction-Oriented Aerial}
 \makebox{Mobility across Heterogeneous Environments}}

\author{Zhibo Duan$^{1}$\thanks{Equal contributions}~~~
\quad Hanlin Kong$^{1}$\footnotemark[1]~~~
\quad Tianran Ma$^{1}$~~~
\quad Yuze Wu$^{1, 2}$~~~
\quad Yuxuan Xu$^{1}$~~~\\
\textbf{Kenghou Hoi}$^{1}$
\hspace{0.65em}\textbf{Changheng Xie}$^{1}$
\hspace{0.65em}\textbf{Xin Zhou}$^{2}$
\hspace{0.65em}\textbf{Jianzhe Gao}$^{1}$
\hspace{0.65em}\textbf{Fei Gao}$^{1, 2}$
\hspace{0.65em}\textbf{Wenguan Wang}$^{1}$\thanks{Corresponding author}
\smallskip
\\
$^1${\small Zhejiang University}
\quad $^2${\small Differential Robotics}
\\
\small \url{https://bozhistudying233.github.io/CrossFly}
}

\crossflyfinalcopy 

\begin{document}
\maketitle
\fancyhead[L]{}
\begin{figure}[H]
  \centering
  \includegraphics[width=1\linewidth]{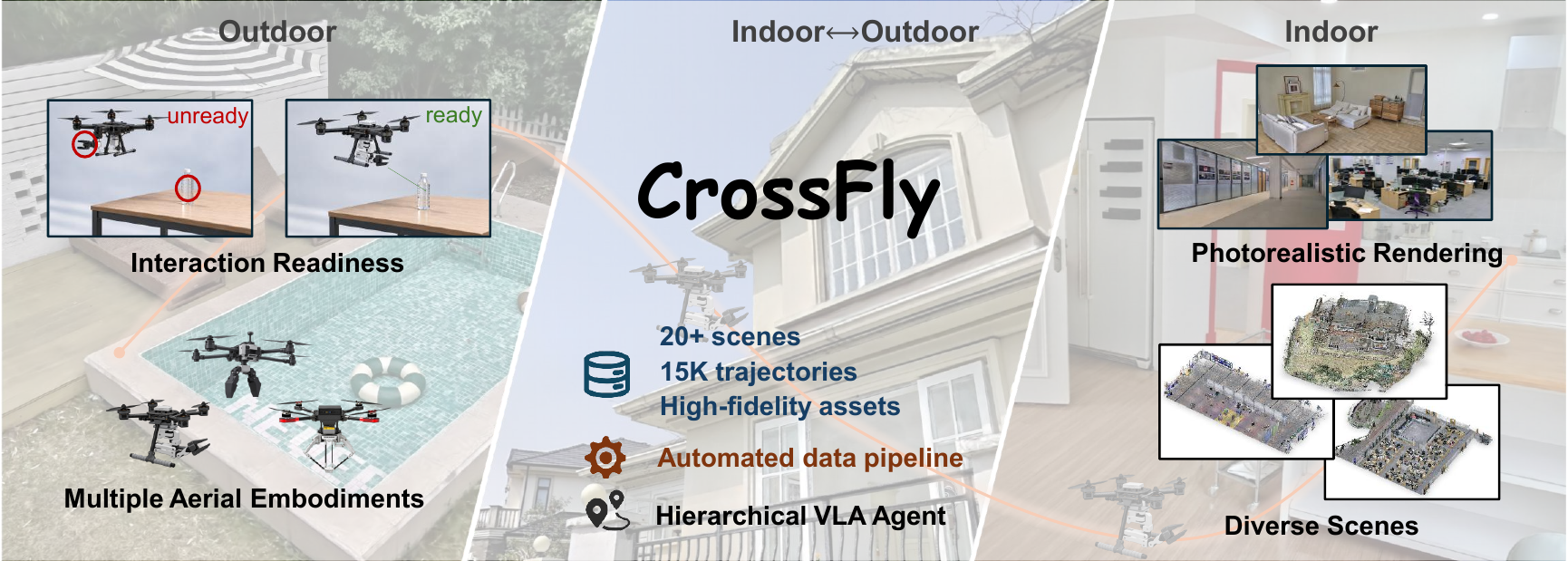}
  \vspace{-20pt}
  \caption{\textbf{Overview of \projectname~benchmark.} It focuses on interaction-oriented aerial mobility tasks and consists of a large dataset for unmanned aerial vehicle (UAV) imitation learning, featuring multiple aerial embodiments, heterogeneous environments, photorealistic rendering, and diverse target objects. To enable systematic experimental analysis under interaction-oriented mobility task settings, we additionally provide an evaluation protocol and propose \mbox{an interaction readiness metric.}}
  \label{fig:overview}
\end{figure}

\begin{abstract}
Aerial mobile manipulation requires unmanned aerial vehicles (UAVs) to traverse heterogeneous environments such as indoor and outdoor domains, and connect navigation with subsequent physical interaction within the same trajectories. Existing aerial benchmarks are typically confined to single-environment settings, limiting evaluation of agents across continuous environmental transitions. When interaction is evaluated, agents are often initialized near the target, leaving the transition from navigation to interaction largely unassessed. This work introduces \projectname, a benchmark for long-horizon interaction-oriented aerial mobility across heterogeneous environments. \projectname~contains \textbf{15K} trajectories across more than \textbf{20} reconstructed scenes under a unified simulation and evaluation framework, covering \textit{Indoor}, \textit{Outdoor}, and \textit{Indoor$\leftrightarrow$Outdoor} trajectories. It further introduces Interaction Readiness Rate (IRR) to assess whether terminal navigation states provide suitable conditions for downstream interaction without assuming a specific manipulation policy. Experiments with \textbf{five} representative aerial vision-language-action (VLA) agents and the proposed \textsc{\projectname-Agent} show that \textit{Indoor$\leftrightarrow$Outdoor} settings are more challenging than single-environment settings, with the highest success rate reaching 9.7\% on unseen scenes. A downstream manipulation case study further validates the relevance of interaction-oriented arrival to subsequent physical interaction. We hope \projectname~helps advance aerial embodied intelligence toward unified mobility and physical interaction in heterogeneous environments.
\end{abstract}

\section{Introduction}
\label{sec:introduction}
Aerial mobile manipulation involves UAVs traversing three-dimensional spaces and interacting with objects within a single trajectory~\citep{airvla, airvla_plus, wang2026ambench, airvla_transfer, he2025flyinghandendeffectorcentricframework}. Reaching an interaction site may require flight through rooms, corridors, open spaces, and building boundaries under varying spatial constraints. Such trajectories may span heterogeneous environments~\citep{airuninav}, where flight must not only reach the target region but also terminate in a state suitable for interaction~\citep{zhang2025momakitchen}. These requirements motivate interaction-oriented aerial mobility as a core capability for extending goal-directed flight toward interaction.

Existing aerial benchmarks are typically confined to single-environment settings, where evaluation centers on instruction-conditioned or goal-directed flight and generalization to unseen scenes~\citep{wang2026uav, OpenFly, aerialvln, citynav, liu2026indooruav}. These protocols largely characterize whether agents reach designated goals, limiting evaluation of their behavior across long-horizon, continuous environmental transitions. When interaction is included, agents are often initialized near the target, providing little assessment of how they progress from extended flight through diverse environments to physical interaction~\citep{airvla, airvla_plus, wang2026ambench}.

Supporting this capability requires continuous trajectories that preserve environmental transitions and the approach to interaction sites~\citep{wu2026navverse}. Such trajectories provide supervision for learning under changing conditions within the same episode, while evaluation should assess whether terminal states support subsequent interaction without assuming a specific manipulation policy.

\projectname~instantiates these requirements as a large-scale benchmark for learning and evaluating interaction-oriented aerial mobility across heterogeneous environments. It contains 15K trajectories across more than 20 reconstructed scenes under a unified simulation framework, spanning diverse \textit{Indoor}, \textit{Outdoor}, and \textit{Indoor$\leftrightarrow$Outdoor} task suites. Each trajectory records synchronized egocentric RGB observations, 6-DoF flight states, language instructions, waypoints, fine-grained subtasks, and interaction-related annotations throughout the trajectory. For evaluation, \projectname~introduces Interaction Readiness Rate (IRR) to assess whether terminal flight states provide suitable conditions for subsequent physical interaction, making it applicable to different manipulation strategies.

Experiments with five representative aerial agents and the proposed \textsc{\projectname-Agent} reveal substantial gaps in interaction-oriented aerial mobility. Even the strongest agent reaches only a 9.7\% success rate on unseen \textit{Indoor$\leftrightarrow$Outdoor} trajectories, indicating difficulty in sustaining goal-directed flight across environmental transitions. IRR shows that goal reaching does not necessarily translate into interaction-oriented arrival, exposing limitations beyond standard navigation metrics. A downstream manipulation case study validates the relevance of interaction readiness to subsequent physical interaction. We hope \projectname~will serve as a standardized platform for evaluating and advancing aerial interaction-oriented mobility.

\section{Related Work}
\label{sec:related-work}
\textbf{Aerial Benchmarks.}
Language-guided aerial benchmarks cover diverse environments and task settings.
AerialVLN~\citep{aerialvln} studies long-horizon UAV instruction following from egocentric observations, while AVDN~\citep{avdn} incorporates interactive dialogue. Subsequent benchmarks improve flight realism and expand real-world scene coverage~\citep{traveluav,citynav}. OpenFly~\citep{OpenFly} integrates data generation, simulation, and agent evaluation across multiple engines, whereas IndoorUAV~\citep{liu2026indooruav} focuses on continuous navigation in indoor environments. Beyond route following, UAV-Flow Colosseo~\citep{wang2026uav} evaluates fine-grained flight control, AIR-VLA~\citep{airvla} couples aerial mobility with manipulation, and HUGE-Bench~\citep{hugebench} introduces process- and safety-aware evaluation for multi-stage tasks. Ground-based last-mile navigation further targets manipulation-ready positioning~\citep{zhang2025momakitchen,lee2026lastmeter}. These benchmarks address complementary aspects of navigation and interaction, but separate indoor and outdoor evaluations do not characterize continuous execution across environmental boundaries. \projectname~brings these dimensions together by combining flight through connected indoor--outdoor environments with terminal interaction-readiness evaluation. Its single-domain and cross-domain suites support systematic comparison of navigation success and arrival quality across different environmental conditions and transitions.

\textbf{Aerial Agents.}
Language-conditioned aerial agents span planning-oriented navigation methods and action-oriented VLA policies. Early navigation approaches predict waypoints or hierarchical trajectories from language and visual observations~\citep{avdn,aerialvln,traveluav}. Foundation-model-based agents support more robust long-horizon decision making through semantic planning, memory, explicit reasoning, and directional grounding~\citep{citynavagent,flightgpt,lookasidevln,htnav,pscavdn,ding2026history}, while AerialVLA~\citep{aerialvla_dialogue} uses task progress and observation history to support proactive dialogue during flight. Action-oriented policies map multimodal observations more directly to flight commands: OpenFly-Agent~\citep{OpenFly} uses historical observations for action prediction, whereas AeroVLA~\citep{aerovla} and AutoFly~\citep{autofly} learn end-to-end continuous control from language instructions. Other work explores learning-free navigation, efficient onboard execution, unified indoor--outdoor training, and compressed long-horizon context~\citep{spf,gradnavpp,airuninav,longfly}. 
Recent aerial manipulation work couples navigation and manipulation over short spatial ranges, leveraging transfer from
manipulation-pretrained VLA models~\citep{airvla_transfer} and separate decoding of flight and manipulator actions~\citep{airvla_plus}. Connecting longer-range navigation to such local interactions requires evaluating whether the terminal flight state supports the subsequent manipulation skill.
\projectname~makes this connection explicit through interaction-readiness evaluation. We further introduce \textsc{\projectname-Agent} as a hierarchical agent, in which a high-level VLM planner specifies the active subtask and a low-level VLA policy generates 6-DoF flight actions and predicts a manipulation-invocation signal.

\begin{figure}[t]
    \centering        
        \includegraphics[
            width=\linewidth
        ]{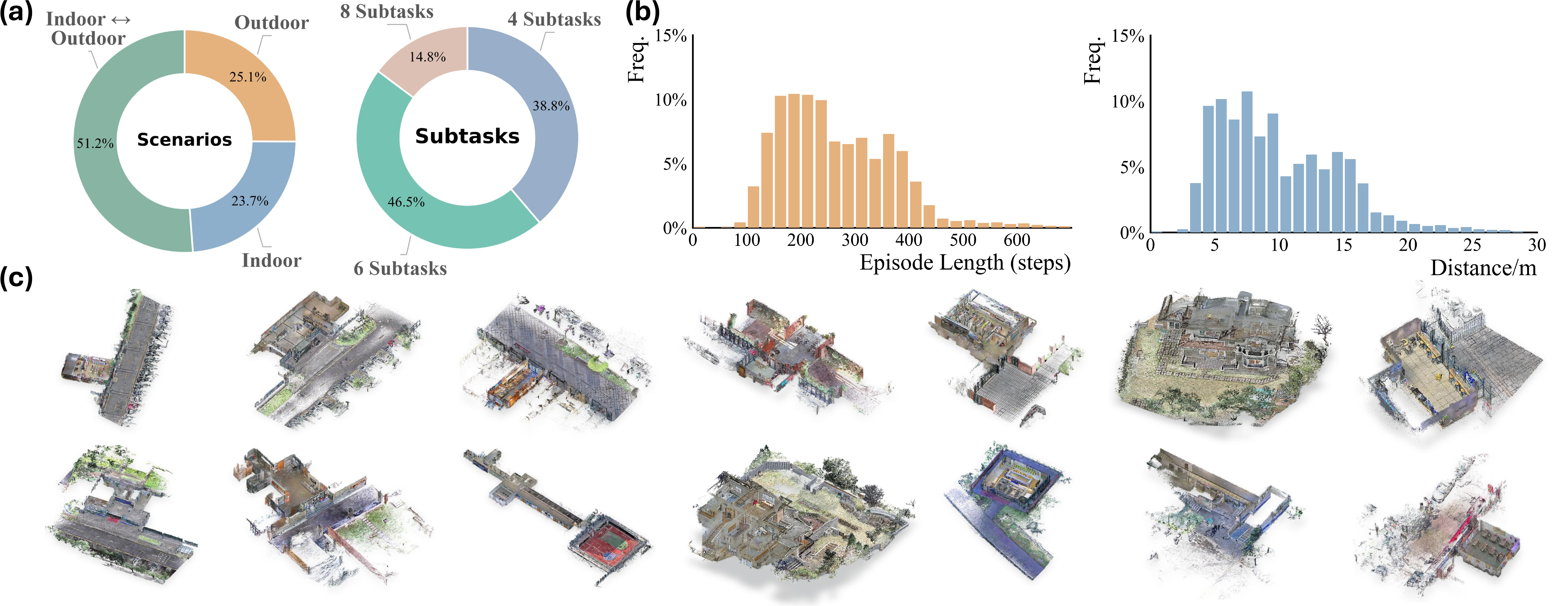}
        \vspace{-20pt}
    \caption{
\textbf{Trajectory statistics and scene diversity.}
(a) Distribution of scene types and task instructions with different numbers of subtasks
(\S\ref{sec:benchmark-stats}).
(b) Distribution of trajectory lengths and path distances, summarizing the temporal and spatial scales of the dataset.
(c) Representative environments in \projectname, where each scene covers indoor and outdoor environments.}

    \label{fig:benchmark_statistics}

\end{figure}

\section{\projectname~Benchmark}
\label{sec:benchmark}

Evaluating interaction-oriented aerial mobility across heterogeneous environments requires unified, consistent definitions of the collected data, execution environment, and evaluation criteria. We first summarize the benchmark composition, including trajectory statistics and scene categories. Then, the dataset construction pipeline describes how scenes,  trajectories, and subtask annotations are constructed (\S\ref{sec:generation-tasks}). Next, the simulation environment specifies the physics simulator, UAV platform, and action space used for closed-loop execution (\S\ref{sec:simulation-environment}). Finally, the evaluation protocol defines the task suites and metrics used to assess \mbox{navigation and interaction readiness (\S\ref{sec:evaluation-protocol}).}

\subsection{Benchmark Stats}
\label{sec:benchmark-stats}
\projectname~contains 15K trajectories with 3M frames, totaling 160+ hours and 120 km of flight.
As shown in Fig.~\ref{fig:benchmark_statistics}, the dataset spans diverse indoor, outdoor, and cross-environment \textit{Indoor$\leftrightarrow$Outdoor} tasks, where cross-environment trajectories account for the largest portion (\textbf{51.2}\%), compared with \textit{Indoor} (\textbf{23.7}\%) and \textit{Outdoor} (\textbf{25.1}\%) trajectories.
The instructions are highly compositional, consisting of 4, 6, or 8 subtasks, with 6-subtask instructions being the most common (\textbf{46.5}\%), followed by 4-subtask (\textbf{38.8}\%) and 8-subtask (\textbf{14.8}\%) instructions.
The trajectories further exhibit substantial variations in both temporal horizon and spatial extent: most trajectories span 100--400 frames, while trajectory distances are broadly distributed across different ranges, with the majority falling below 20\,m and a small number extending to nearly 30\,m.
Such diversity in trajectory length and spatial scale introduces varying levels of long-horizon navigation and interaction complexity.
Moreover, \projectname~covers a wide range of heterogeneous environments with substantially different layouts, scales, and visual appearances, providing diverse settings for evaluating embodied aerial agents.

As shown in Table$_{\!}$~\ref{tab:benchmark-comparison}, \projectname~presents a more challenging setting compared with existing benchmarks. The agent needs to navigate across heterogeneous environments with different spatial constraints, while ensuring that the destinations are suitable for subsequent physical interactions.

\subsection{Dataset Construction Pipeline}
\label{sec:generation-tasks}
\begin{figure}[t]
  \centering
  \includegraphics[width=\linewidth]{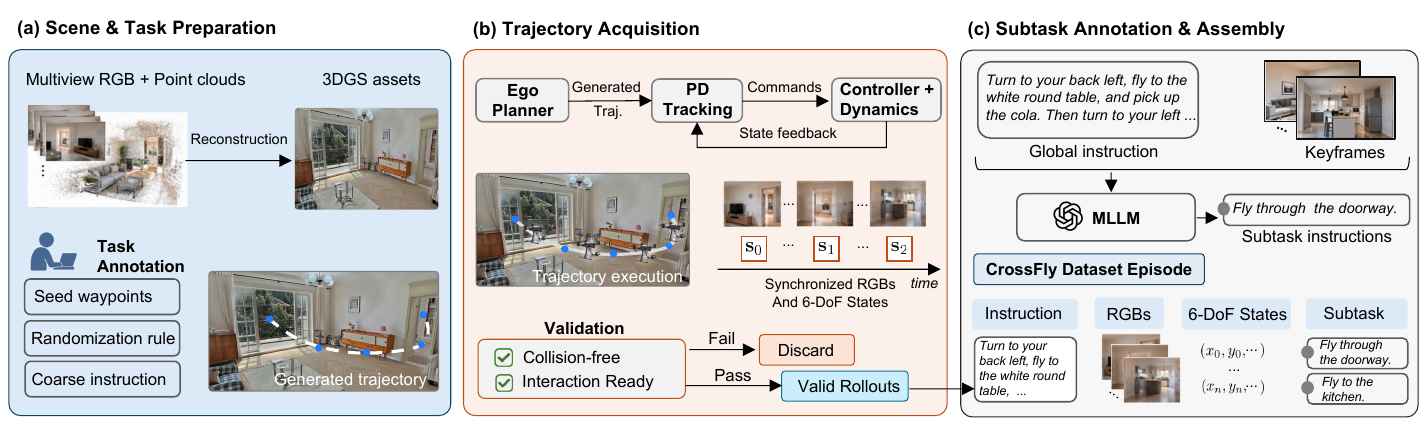}
  \vspace{-20pt}
  \caption{\textbf{Overview of the dataset construction pipeline (\S\ref{sec:generation-tasks}).} \projectname~begins with reconstructed scenes from point clouds and realistic rendering, followed by waypoint annotation and trajectory acquisition through UAV simulation. The collected trajectories are then assembled with RGB observations, UAV states, and task-related annotations to construct the benchmark dataset.}
  \label{fig:data-pipeline}
\end{figure}

This pipeline is designed to construct large-scale, executable, and semantically annotated UAV trajectories in a unified framework across diverse heterogeneous environments, including indoor, outdoor, and indoor--outdoor domains. As shown in Fig.~\ref{fig:data-pipeline}, it consists of three main stages. First, tasks are defined through objects, key waypoints, and language instructions in self-collected scenes. Next, the resulting task specifications are executed under realistic UAV dynamics to acquire synchronized egocentric observations and flight states, while infeasible rollouts are filtered out automatically. Finally, the waypoint structure and trajectory observations are used to derive fine-grained task annotations, providing structured supervision at both the trajectory and subtask level for effective learning.

\begin{table*}[!t]
  \setlength{\abovecaptionskip}{0cm}
  \setlength{\belowcaptionskip}{6pt}
  \centering
  \caption{\textbf{Comparison with related benchmarks (\S\ref{sec:benchmark-stats}).}}
  \label{tab:benchmark-comparison}
  \resizebox{\textwidth}{!}{%
    \setlength\tabcolsep{4pt}
    \renewcommand\arraystretch{1.02}
    \begin{tabular}{r||cccccccc}
      \hline\thickhline
      \rowcolor{mygray}
      \multicolumn{1}{c||}{Dataset}
      & Platform
      & Scene
      & $N_{\text{traj}}$
      & $N_{\text{scenes}}$
      & Environment
      & Interaction
      & Dynamics
      & DoF\\
      \hline\hline

      R2R~\citep{anderson2018vision}
        & Ground Agent & Matterport3D & 7,189 & 90
        & Indoor & \xmark & \xmark & Graph-based\\

      VLN-CE~\citep{krantz2020beyond}
        & Ground Agent & Habitat & 4,475 & 90
        & Indoor & \xmark & \xmark & 2\\

      OpenUAV~\citep{schmittle2018openuav}
        & UAV & AirSim + UE & 12,149 & 22
        & Outdoor & \xmark & \cmark & 6\\
      AVDN~\citep{avdn}
        & UAV & xView & 6,269 & --
        & Outdoor & \xmark & \xmark & 3\\

      CityNav~\citep{citynav}
        & UAV & SensatUrban & 32,637 & 34
        & Outdoor & \xmark & \xmark & 4\\

      OpenFly~\citep{OpenFly}
        & UAV & AirSim, GTAV, etc. & 100K & 18
        & Outdoor & \ding{55} & \xmark & 4\\

      IndoorUAV~\citep{liu2026indooruav}
        & UAV & Habitat & 16,040 & 1,075
        & Indoor & \xmark & \cmark & 4\\

      \hline

      \multicolumn{1}{c||}{\projectname}
        & UAV & Isaac Sim & 15,263 & 23
        & Heterogeneous & \cmark & \cmark & 6\\

      \hline
    \end{tabular}}
  \vspace{-8pt}
\end{table*}

\textbf{Scene Collection.}
We reconstruct diverse heterogeneous environments from multi-view RGB images and point clouds captured by a handheld 3D scanner, including 3D Gaussian Splatting (3DGS) scenes and mesh representations~\citep{kerbl3Dgaussians}. The collected scenes cover both individual indoor and outdoor spaces as well as connected regions that allow trajectories to transition between the two domains. These reconstructed environments provide the visual appearance and scene geometry used for subsequent trajectory construction and simulation.
For each scene, we annotate target objects, spatial waypoints, and language instructions to define executable trajectories. Each trajectory is specified by four kinds of waypoints: \textit{start}, \textit{intermediate}, \textit{terminal}, and \textit{turning} points. By combining these annotations with different target objects and language instructions, each reconstructed scene can support diverse configurations with different routes, viewpoints, and interaction~goals.

\textbf{Waypoint Annotation.}
Within each reconstructed scene, we define a high-level trajectory by annotating target objects, language instructions, and a sparse set of key waypoints. The waypoints serve different roles along the path: the \textit{start} waypoint specifies the initial UAV pose; \textit{intermediate} waypoints guide UAV through important spatial regions along the intended route; the \textit{terminal} waypoint specifies the final destination, typically near the target object. Together, these annotations define a coarse trajectory representation for subsequent trajectory refinement.

\textbf{Trajectory Acquisition.}
Given the annotated waypoints, UAV is initialized at the \textit{start} pose. Then the Ego-Planner~\citep{zhou2022swarm} generates a collision-free path between consecutive waypoints, transforming the sparse waypoints into an executable reference trajectory. A position-velocity PD tracking controller then converts the trajectory into acceleration commands executed by the low-level flight controller. The resulting motion is simulated under UAV dynamics, introducing realistic flight effects such as small oscillations, tracking deviations, response delays, and viewpoint~perturbations.

During each rollout, we record synchronized egocentric RGB observations and 6-DoF UAV states. The complete state sequence provides a temporally aligned representation of how the agent progresses through the task and supports fine-grained evaluation of navigation and interaction behaviors. Rollouts that collide with the reconstructed scene geometry or fail to reach the designated waypoint are automatically rejected. This execution-and-filtering procedure ensures that retained task instances are physically feasible and can be completed under the simulator dynamics.

\textbf{Task Decomposition.}~
The waypoint structure provides a natural basis for partitioning each trajectory into semantically coherent subtasks. Each trajectory is partitioned into a sequence of subtasks, where each subtask corresponds to the trajectory segment between two adjacent waypoints. These waypoints include \textit{start}, \textit{intermediate}, and \textit{terminal} waypoints. For each segment, we extract the egocentric observations at its start and end frames together with the global instruction $\ell$, and prompt a pretrained VLM (GPT-5.5) to generate a concise description of the segment-level intent, which serves as the ground-truth subtask for training high-level planning models after human verification.

\subsection{Simulation Environment}
\label{sec:simulation-environment}

\textbf{Physical Simulator.}~\projectname~is built on NVIDIA Isaac Sim, which provides physics simulation, collision detection, and closed-loop UAV control. UAV dynamics are simulated with PhysX, and the reconstructed 3D Gaussian Splatting scenes serve as the visual environments for navigation. At each simulation step, egocentric observations are rendered according to the current UAV pose, while the predicted actions are executed under the simulator dynamics. The updated UAV state is subsequently used to generate the next observation, forming a closed-loop interaction cycle between the agent and the simulated environment. This setup enables closed-loop evaluation across indoor, outdoor, and heterogeneous environments with realistic visual appearance and physical motion.

\textbf{UAV Platform.}
\projectname~supports diverse aerial embodiments through a unified control interface. Our main benchmark experiments are conducted on two mobile manipulation platforms: a quadrotor equipped with SO100-ARM~\citep{cadene2024lerobot} and a hand-like UAV~\citep{Wu_2026}.

\textbf{Action Space.}
We adopt a 6-DoF action space for UAV navigation, consisting of position $(x,y,z)$ and orientation (roll, pitch, and yaw). Motion commands predicted by the agent are executed through a low-level PD controller. For object-interaction tasks, agents can additionally issue an interaction command to invoke the corresponding manipulation policy after reaching the target.

\subsection{Evaluation Protocol}
\label{sec:evaluation-protocol}
\textbf{Task Suites.}
While many existing benchmarks focus on either indoor or outdoor domains, \projectname~provides a cross-domain dataset with aerial interaction-oriented mobility scenarios spanning heterogeneous environments. It provides three fixed suites for evaluation: \textit{\textbf{Indoor}}, \textit{\textbf{Outdoor}}, and \textit{\textbf{Indoor$\leftrightarrow$Outdoor}}. The first two measure navigation and interaction within a single environmental domain, and the last one evaluates long-horizon execution across environmental transitions.

\begin{itemize}[leftmargin=*]
	\setlength{\itemsep}{0pt}
	\setlength{\parsep}{-0pt}
	\setlength{\parskip}{-0pt}
	\setlength{\leftmargin}{-8pt}
\item \textbf{\textit{Indoor.}}
Indoor tasks are executed inside buildings. Given a natural-language instruction, the agent must navigate through rooms, corridors, and doorways, approach the specified interaction location, and output a special token such as \texttt{<Manip>} when manipulation is required. The suite evaluates instruction following, object grounding, and target approach in constrained indoor spaces.

\item \textbf{\textit{Outdoor.}}
Outdoor tasks follow the same formulation in courtyards, building surroundings, and other open areas. They emphasize long-range navigation, instruction grounding, and target localization under larger spatial scales and different visual and geometric conditions.

\item \textbf{\textit{Indoor$\leftrightarrow$Outdoor.}}
These tasks require the agent to navigate across indoor and outdoor environments in either direction, starting in one environment and reaching a target located in the other. The agent must identify and approach an appropriate transition region, such as a doorway or building entrance, cross the environmental boundary, and continue toward the target. This suite evaluates robust navigation under substantial changes in appearance, illumination, and spatial constraints, while maintaining effective navigation performance across heterogeneous environments.

\end{itemize}

\textbf{Interaction Readiness Rate.}
Alongside standard navigation metrics, we report the proposed
Interaction Readiness Rate (IRR) to evaluate whether navigation terminates
in a state suitable for subsequent physical interaction.
Let $\mathcal{R}$ denote the set of episodes in which UAV successfully
reaches the target interaction region. 
We define the interaction-ready
terminal state set as:
\begin{equation}
\mathcal{X}_{\mathrm{int}}
=
\left\{
s \;\middle|\;
d(s) \leq d_{\max},\;
\psi(s) \in [\psi_{\min},\psi_{\max}]
\right\},
\end{equation}
where $d(s)$ denotes the distance from the terminal UAV state to the interaction point, and $\psi(s)$ denotes the relative yaw offset between UAV
heading and the target-facing direction.
IRR is then defined as:
\begin{equation}
\mathrm{IRR}
=
\begin{cases}
\frac{1}{|\mathcal{R}|}
\sum\nolimits_{i \in \mathcal{R}}
\mathbb{I}\!\left[s_i^{T} \in \mathcal{X}_{\mathrm{int}}\right],
& |\mathcal{R}|>0,\\[6pt]
0,
& |\mathcal{R}|=0.
\end{cases}
\end{equation}
Here $s_i^{T}$ denotes the terminal state of trajectory $i$ and
$\mathbb{I}[\cdot]$ is the indicator function. 
We set
$d_{\max}=0.626\,\mathrm{m}$ and
$[\psi_{\min},\psi_{\max}]
=
[-50.5^\circ,47.5^\circ]$.
These thresholds define the required spatial proximity and viewpoint alignment
for downstream interaction, and are further analyzed in \S\ref{sec:irr-analysis}.

IRR therefore measures the fraction of target-reaching trajectories whose
terminal states satisfy the defined interaction-ready conditions, thereby
quantifying terminal-state suitability beyond navigation completion.
Conditioning on successful target reaching isolates terminal-state
quality from navigation success, making IRR complementary to standard
navigation metrics such as SR.

\begin{figure}[t]
  \centering
  \includegraphics[width=\linewidth]{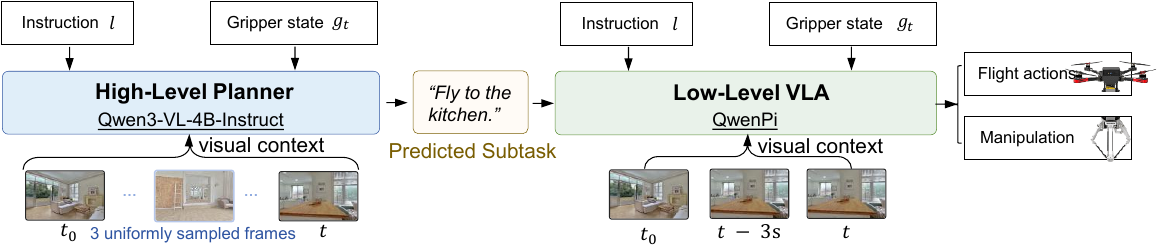}
  \vspace{-20pt}
  \caption{\textbf{Architecture of \textsc{\projectname-Agent~(\S\ref{sec:setup})}.} A high-level VLM planner uses visual context spanning the observed trajectory to predict the active language subtask. Conditioned on this subtask, a low-level VLA policy combines the initial observation with recent visual context to generate flight action chunks and determine when to predict a manipulation-invocation signal.}
  \label{fig:hierarchical-baseline}
\end{figure}
\section{Experiment}
\projectname~is proposed to support long-horizon aerial interaction-oriented mobility across heterogeneous environments with diverse task settings. To demonstrate its broad applicability and evaluate the capabilities of existing methods, extensive experiments are conducted. First, the evaluation metrics and baseline methods are introduced in \S\ref{sec:setup}. Second, representative methods are benchmarked on seen scenes and unseen scenes in \S\ref{sec:performance}. Then, qualitative analyses of representative trajectories are presented in \S\ref{sec:qualitative}. Finally, we conduct diagnostic experiments to analyze the effects of environmental transitions and examine the connection between IRR and manipulation in \S\ref{sec:diagnostic-experiments}.

\subsection{Experimental Setup}
\label{sec:setup}

\textbf{Evaluation Metric.} Following standardized protocols~\citep{wang2026uav, liu2026indooruav}, we use three standard metrics for evaluation: i) Success Rate (SR); ii) Navigation Error (NE); and iii) Oracle Success Rate (OSR). In addition, the proposed Interaction Readiness Rate (IRR) is reported to evaluate whether UAV reaches a state suitable for interaction after completing navigation tasks.

\textbf{Baselines.}
We evaluate several representative open-source agents spanning different action-generation paradigms and training distributions. $\pi_0$~\citep{black2024pi_0} is a Flow Matching-based foundation model pretrained on large-scale cross-embodiment robot data. $\pi_0$-FAST~\citep{pertsch2025fastefficientactiontokenization} uses FAST action tokenization for efficient autoregressive action generation. OpenFly-Agent~\citep{OpenFly} represents agents developed for outdoor aerial scenarios, OpenVLA-OFT~\citep{kim2025fine} represents pretrained VLA agents adapted to downstream control tasks through optimized fine-tuning, and GR00T-N1.7~\citep{bjorck2025gr00t} represents generalist VLA agents developed for humanoid robot control. For a fair comparison, all baselines are retrained on \projectname~training datasets following unified training recipes. More details are provided in \S\ref{app:implementation}.

\textbf{\projectname-Agent.}
We construct \textsc{\projectname-Agent} as a hierarchical agent for long-horizon interaction-oriented aerial mobility (Fig.~\ref{fig:hierarchical-baseline}), consisting of a high-level VLM planner and a low-level VLA policy.
The planner predicts the subtask from the instruction, gripper state, and visual context comprising the initial frame, three uniformly sampled historical frames, and the current frame.
The policy uses the initial frame, a frame from 3~s earlier, and the current frame.
Conditioned on this visual context, the predicted subtask, global instruction, and gripper state, the policy adopts QwenPi~\citep{community2026starvla} to generate action chunks together with a manipulation-invocation signal.
The planner and policy are trained independently and coupled through the language subtask interface.
The planner is fine-tuned on annotated subtasks, while the policy learns action chunks via flow matching, with supervision for manipulation invocation.
The policy uses ground-truth subtasks during training and planner predictions at inference. See \S\ref{appendix:method} for more details.

\subsection{Performance Benchmarking}  
\label{sec:performance}  
\textbf{Performance on Seen Scenes.}  
Table~\ref{tab:seen-results} reports the performance of different methods on seen scenes.  
These results characterize the capability of existing state-of-the-art methods in environments included in the training distribution.  
\textsc{\projectname-Agent} achieves the best performance across all metrics and task suites, indicating that explicit high-level task decomposition is useful for interaction-oriented aerial mobility tasks.  
Compared with existing methods without high-level planning~\citep{black2024pi_0, pertsch2025fastefficientactiontokenization, OpenFly, kim2025fine, bjorck2025gr00t}, \textsc{\projectname-Agent} obtains consistent improvements across different task suites.  
For example, on the \textit{Indoor} suite, it achieves SR of \textbf{23.5\%} compared to 17.7\% from $\pi_0$~\citep{black2024pi_0}, while also improving IRR from 37.2\% to \textbf{51.0\%}. These results show that \projectname~evaluates both target reaching and interaction readiness, revealing performance differences among aerial navigation agents.  
  
\begingroup  
\setlength{\intextsep}{6pt}  
\begin{table}[!t]  
\setlength{\abovecaptionskip}{0cm}  
\setlength{\belowcaptionskip}{6pt}  
\centering  
\caption{\textbf{Quantitative comparison on seen scenes (\S\ref{sec:performance}).}}  
\label{tab:seen-results}  
\resizebox{\textwidth}{!}{%
\setlength\tabcolsep{4pt}  
\renewcommand\arraystretch{1.02}  
\begin{tabular}{r||cccc|cccc|cccc}  
\hline\thickhline  
  
\rowcolor{mygray}  
{}  
& \multicolumn{4}{c|}{\textit{Indoor}}  
& \multicolumn{4}{c|}{\textit{Outdoor}}  
& \multicolumn{4}{c}{\textit{Indoor$\leftrightarrow$ Outdoor}} \\  
\cline{2-13}  
  
\rowcolor{mygray}  
\multicolumn{1}{c||}{\multirow{-2}{*}{Model}}  
& NE $\downarrow$ & SR $\uparrow$ & OSR $\uparrow$ & IRR $\uparrow$  
& NE $\downarrow$ & SR $\uparrow$ & OSR $\uparrow$ & IRR $\uparrow$  
& NE $\downarrow$ & SR $\uparrow$ & OSR $\uparrow$ & IRR $\uparrow$ \\  
\hline\hline  
  
$\pi_0$~\citep{black2024pi_0}  
& 4.07 & 17.7 & 23.6 & 37.2  
& 3.79 & 20.4 & 27.8 & 37.0  
& 7.21 & 13.2 & 19.7 & 29.1 \\  
  
$\pi_0$-FAST~\citep{pertsch2025fastefficientactiontokenization}  
& 3.81 & 18.5 & 19.2 & 39.0  
& 3.42 & 22.7 & 30.1 & 38.9  
& 6.63 & 15.8 & 22.1 & 30.7 \\  
  
OpenFly-Agent~\citep{OpenFly}  
& 3.48 & 15.1 & 21.4 & 46.7  
& 3.39 & 18.1 & 20.6 & 43.4  
& 5.93 & 13.9 & 17.7 & 34.2 \\  
  
OpenVLA-OFT~\citep{kim2025fine}  
& 3.84 & 17.2 & 21.7 & 47.9  
& 3.51 & 19.5 & 24.4 & 38.1  
& 6.87 & 14.9 & 21.3 & 38.8 \\  
  
GR00T-N1.7~\citep{bjorck2025gr00t}  
& 3.66 & 18.9 & 19.7 & 40.2  
& 3.43 & 21.3 & 24.2 & 40.0  
& 6.37 & 11.7 & 15.6 & 31.5 \\  
  
\hline  
  
\multicolumn{1}{c||}{\textbf{Ours}}  
& \textbf{3.32} & \textbf{23.5} & \textbf{30.5} & \textbf{51.0}
& \textbf{3.17} & \textbf{25.5} & \textbf{31.7} & \textbf{50.3}
& \textbf{5.42} & \textbf{19.8} & \textbf{24.3} & \textbf{43.0} \\

\multicolumn{1}{c||}{Error Bar ($\pm$)}  
& (0.11) & (1.6) & (1.7) & (1.1)  
& (0.10) & (1.5) & (1.6) & (1.0)  
& (0.15) & (1.8) & (1.9) & (1.2) \\  
\hline  
  
\hline  
\end{tabular}}  
\end{table}  
\endgroup  
  
\begin{table}[!t]  
\setlength{\abovecaptionskip}{0cm}  
\setlength{\belowcaptionskip}{6pt}  
\centering  
\caption{\textbf{Quantitative comparison on unseen scenes (\S\ref{sec:performance}).}}  
\label{tab:unseen-results}  
\resizebox{\textwidth}{!}{%
\setlength\tabcolsep{4pt}  
\renewcommand\arraystretch{1.02}  
\begin{tabular}{r||cccc|cccc|cccc}  
\hline\thickhline  
  
\rowcolor{mygray}  
{}  
& \multicolumn{4}{c|}{\textit{Indoor}}  
& \multicolumn{4}{c|}{\textit{Outdoor}}  
& \multicolumn{4}{c}{\textit{Indoor$\leftrightarrow$ Outdoor}} \\  
\cline{2-13}  
  
\rowcolor{mygray}  
\multicolumn{1}{c||}{\multirow{-2}{*}{Model}}  
& NE $\downarrow$ & SR $\uparrow$ & OSR $\uparrow$ & IRR $\uparrow$  
& NE $\downarrow$ & SR $\uparrow$ & OSR $\uparrow$ & IRR $\uparrow$  
& NE $\downarrow$ & SR $\uparrow$ & OSR $\uparrow$ & IRR $\uparrow$ \\  
\hline\hline  
  
$\pi_0$~\citep{black2024pi_0}  
& 5.72 & 10.8 & 13.4 & 26.7  
& 5.26 & 13.5 & 15.8 & 27.8  
& 9.38 & 2.4 & 4.1 & 21.7 \\  
  
$\pi_0$-FAST~\citep{pertsch2025fastefficientactiontokenization}  
& 5.18 & 13.2 & 15.7 & 28.4  
& 4.83 & 15.6 & 17.2 & 29.1  
& 8.74 & 3.7 & 6.3 & 25.2 \\  
  
OpenFly-Agent~\citep{OpenFly}  
& 4.49 & 10.6 & 13.9 & 32.7  
& 4.22 & 13.4 & 15.1 & 26.4  
& 7.61 & 7.2 & 7.9 & 26.2 \\  
  
OpenVLA-OFT~\citep{kim2025fine}  
& 5.45 & 11.9 & 14.6 & 27.5  
& 4.97 & 14.1 & 16.4 & 28.2  
& 9.02 & 3.1 & 4.7 & 27.9 \\  
  
GR00T-N1.7~\citep{bjorck2025gr00t}  
& 5.03 & 13.8 & 16.2 & 28.8  
& 4.61 & 16.2 & 18.0 & 29.7  
& 8.56 & 4.2 & 5.6 & 23.8 \\  
  
\hline  
  
\multicolumn{1}{c||}{\textbf{Ours}}  
& \textbf{4.05} & \textbf{17.3} & \textbf{18.6} & \textbf{35.5}
& \textbf{3.83} & \textbf{18.5} & \textbf{22.7} & \textbf{35.3}
& \textbf{6.91} & \textbf{9.7} & \textbf{11.4} & \textbf{29.3} \\

\multicolumn{1}{c||}{Error Bar ($\pm$)}  
& (0.17) & (1.4) & (1.5) & (1.1)  
& (0.16) & (1.5) & (1.5) & (1.1)  
& (0.24) & (0.9) & (1.1) & (0.7) \\  
  
\hline  
\end{tabular}}  
\end{table}

\textbf{Generalization on Unseen Scenes.}
As shown in Table~\ref{tab:unseen-results}, all agents experience performance degradation in unseen environments, while \textsc{\projectname-Agent} maintains the best performance across all metrics and task suites.
In particular, on the \textit{Outdoor} task suite, \textsc{\projectname-Agent} improves SR from 16.2\% to \textbf{18.5\%}, and reduces NE from 4.22 to \textbf{3.83}.
These results suggest that hierarchical planning and subtask supervision in \textsc{\projectname-Agent} provide useful guidance in unseen scenes.

\subsection{Qualitative Results}
\label{sec:qualitative}
\begin{figure}[!t]
  \centering
  \includegraphics[width=\linewidth]{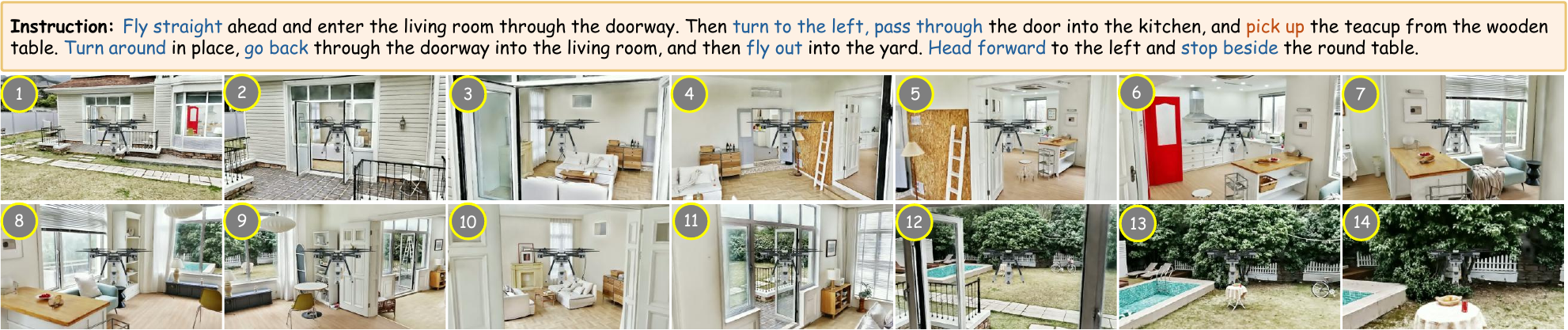}
  \vspace{-20pt}
  \caption{\textbf{Qualitative results of \textsc{\projectname-Agent}~(\S\ref{sec:qualitative}).} This example shows the instruction and selected third-person views. Blue text denotes navigation actions, while red text denotes manipulation actions. The top row shows the trajectory from the start to the interaction-ready state, while the bottom row covers the trajectory from the interaction-ready state to task completion.}
  \label{fig:qualitative-results}
\end{figure}

Fig.~\ref{fig:qualitative-results} presents the qualitative results of \textsc{\projectname-Agent}. The example illustrates a challenging \textit{Indoor$\leftrightarrow$Outdoor} task, requiring the agent to adapt its navigation behavior when transitioning between indoor and outdoor environments. More visualizations are provided in \S\ref{appendix:qualitative}.
These qualitative results further demonstrate the effectiveness of the hierarchical design and subtask supervision in handling navigation tasks with varying scene constraints and planning horizons.

\subsection{Diagnostic Experiments}
\label{sec:diagnostic-experiments}
\textbf{Controlled Comparison across Task Suites.}
To investigate whether the performance degradation of \textit{Indoor$\leftrightarrow$Outdoor} tasks is associated with environmental transitions rather than differences in other trajectory properties, we conduct a controlled comparison with matched trajectory statistics. Specifically, we construct matched subsets for the \textit{Indoor} and \textit{Outdoor} suites and the \textit{Indoor$\leftrightarrow$Outdoor} suite by controlling for trajectory length, number of subtasks, and instruction length. We select 5 reconstructed scenes and sample 30 matched trajectories from each suite within each scene. This design minimizes the influence of variations in navigation horizon, scene gaps, and language instruction variations, enabling a more controlled comparison across the three task suites.

\setlength\intextsep{0pt}
\begin{wraptable}{r}{0.66\linewidth}
  \setlength{\abovecaptionskip}{0pt}
  \caption{\textbf{Performance comparison under matched trajectories.}}
  \resizebox{0.66\textwidth}{!}{
  \setlength\tabcolsep{3pt}
  \renewcommand\arraystretch{1.05}
  \begin{tabular}{c|ccc|cc}
      \hline\thickhline
      \rowcolor{mygray}
      Task Suite
      & Traj. Length
      & \# Subtasks
      & Instr. Length
      & NE $\downarrow$
      & SR $\uparrow$
      \\
      \hline\hline

      \textit{Indoor}
      & 11.2
      & 6.8
      & 18.9
      & 4.15
      & 23.64
      \\

      \textit{Outdoor}
      & 11.8
      & 6.7
      & 18.5
      & 4.36
      & 25.28
      \\

      \hline

      \textit{Indoor$\leftrightarrow$Outdoor}
      & 11.5
      & 6.8
      & 18.8
      & 6.43
      & 10.98
      \\

      \hline
  \end{tabular}
  }
  \label{tab:transition-effect}
\end{wraptable}

As shown in Table~\ref{tab:transition-effect}, all task suites contain comparable trajectories. Despite receiving similar training supervision, agents consistently achieve lower performance on the \textit{Indoor$\leftrightarrow$Outdoor} task suite compared with single-domain tasks. These results suggest that the performance gap cannot be fully explained by differences in training exposure or trajectory properties. Instead, cross-domain navigation presents additional challenges that require agents to adapt to changes in visual appearance and spatial constraints when transitioning between different domains.

\begin{figure}[!t]
  \centering
  \includegraphics[width=\linewidth]{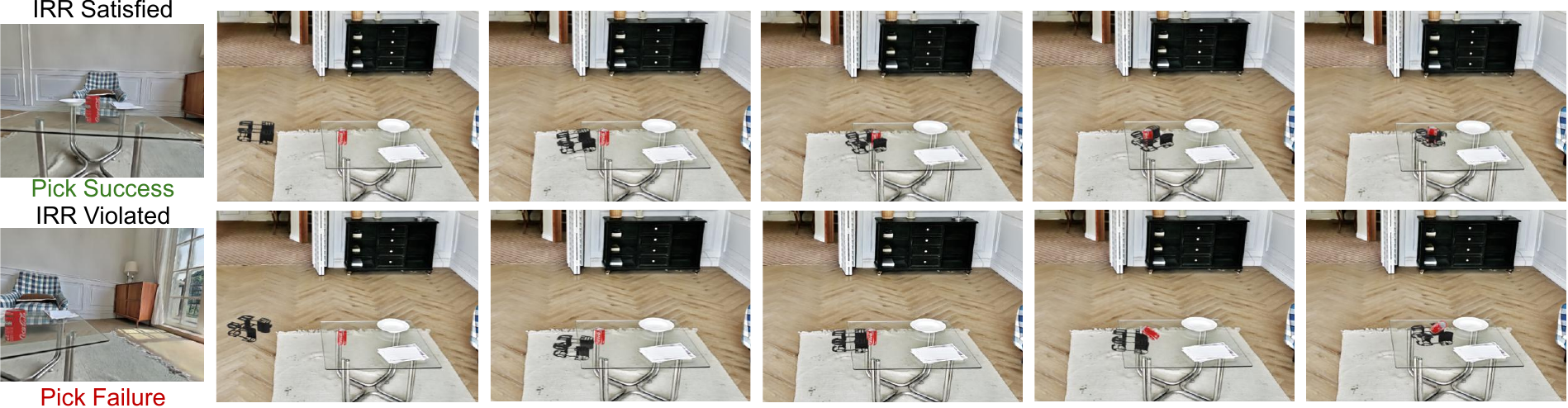}
  \vspace{-20pt}
  \caption{\textbf{Manipulation case study~(\S\ref{sec:case-study}).} Each row represents a manipulation case. The leftmost image shows the initial observation of UAV. The top case satisfies IRR criteria and achieves a successful grasp, whereas the bottom case violates IRR criteria and fails to grasp.}
  \label{fig:manipulation-case}
\end{figure}

\textbf{Manipulation Case Study.}
\label{sec:case-study}
To examine whether IRR reflects the suitability of a navigation terminal state for subsequent manipulation, we train a reinforcement learning policy for grasping on a hand-like UAV platform, using the target object pose estimated from the depth map as input. The grasping policy is initialized from the terminal state produced by the navigation agent. As shown in Fig.~\ref{fig:manipulation-case}, the top case satisfies the distance and target-view constraints defined by IRR, enabling the policy to successfully grasp the target object. In contrast, the bottom case violates IRR criteria, and the same policy fails to complete the grasp. These examples demonstrate that IRR can distinguish navigation terminal states that lead to different outcomes \mbox{under the same downstream grasping policy.}

\textbf{Empirical Validation of IRR Criteria.}
\label{sec:irr-analysis}
To ground IRR criteria in downstream manipulation capability, we evaluate manipulation success across multiple aerial embodiments, each with its own manipulation policy.
For each embodiment, we randomly sample terminal UAV states around fixed target objects and record grasp outcomes while varying the distance to the target and relative yaw.
We then estimate the interaction readiness boundaries through maximum-likelihood fitting of the grasp outcomes, using candidate spatial and viewpoint partitions induced by the sampled terminal states.
Aggregating the manipulation outcomes across all embodiments and their corresponding policies yields the spatial threshold $\hat d_{\max}=0.626$ m and the valid viewpoint interval $[\hat\psi_{\min},\hat\psi_{\max}]=[-50.5^\circ,47.5^\circ]$.
The resulting region provides an empirical approximation of the terminal configurations that support downstream grasping.
Thus, IRR criteria are grounded in manipulation feasibility across multiple embodiments rather than manually specified thresholds.

\section{Limitations and Impact}
\textbf{Limitations.} Despite the effectiveness of \projectname~in providing a unified evaluation framework for interaction-oriented aerial mobility across heterogeneous environments, it still has several limitations that could be addressed in future work.
\textbf{First}, \projectname~is currently developed in simulation environments. While the benchmark incorporates realistic rendering and physics-based UAV dynamics, the simulation-to-real gap caused by sensor noise, aerodynamic disturbances, and hardware constraints remains unexplored. Future work will investigate real-world deployment and transfer protocols.
\textbf{Second}, \projectname~evaluates agents under feasible trajectories generated from successful demonstrations, while real-world long-horizon autonomy requires robust failure recovery and online adaptation. Future benchmarks could incorporate interactive learning, recovery behaviors, and continual adaptation to better evaluate embodied intelligence in open-world environments.

\textbf{Social Impact.} \projectname~advances the evaluation of interaction-oriented aerial mobility by assessing goal reaching and whether terminal states are suitable for interaction across heterogeneous environments. On the positive side, it may facilitate the development of aerial embodied agents. For potential negative impact, \projectname~ is simulation-based and does not fully capture hardware constraints or failure recovery. Hence, it requires further safety validation~\citep{feng2026bits}.

\section{Conclusion}\label{sec:conclusion}
In this work, we introduce \projectname, a benchmark for interaction-oriented aerial mobility across heterogeneous environments. It provides 15K trajectories across 23 reconstructed scenes and evaluates both navigation performance and interaction readiness through a unified protocol. We propose the Interaction Readiness Rate to measure whether the terminal state satisfies the spatial and viewpoint requirements for interaction. Experiments with five aerial agents and \textsc{\projectname-Agent} show that existing methods face challenges in indoor--outdoor transition tasks, highlighting the difficulty of connecting long-horizon navigation with physical interaction. We hope \projectname~facilitates the development of interaction-oriented aerial mobility by providing a standardized benchmark.
\newpage
\bibliography{references}
\bibliographystyle{unsrtnat}

\newpage
\appendix
In the appendix, we provide the following items that shed deeper insight into our contributions:
\begin{itemize}[leftmargin=*, itemsep=0pt, topsep=2pt]
\item \S\ref{appendix:method}: Additional details on the \textsc{\projectname-Agent}.
\item \S\ref{app:benchmark_split}: Supplementary information of benchmark splits.
\item \S\ref{app:implementation}: Implementation details.
\item \S\ref{Discussion}: Discussion.
\end{itemize}

\section{Additional details on the \textsc{\projectname-Agent}}
\label{appendix:method}
\subsection{Method}
\textbf{Overview.}
We introduce \textsc{\projectname-Agent}, a hierarchical agent for aerial mobile manipulation.
A high-level VLM planner predicts language subtasks using visual observations sampled across the trajectory.
Conditioned on the predicted subtask, a low-level VLA policy uses the initial and recent observations to generate flight action chunks and determine when to invoke manipulation.

\textbf{High-Level Planning.}
Let $\ell$ denote the global instruction, $o_t$ the current RGB observation, and $g_t$ the gripper state.
The high-level VLM planner uses Qwen3-VL-4B-Instruct~\citep{qwen3technicalreport} to predict the active language subtask.
Its visual context consists of the initial observation $o_0$, three historical frames $\mathcal{R}^{\mathrm{plan}}_t$ uniformly sampled between the initial and current observations, and the current observation $o_t$:
\begin{equation}
\label{eq:subtask-planning}
z_t=P_\theta (\ell,g_t,o_0,\mathcal{R}^{\mathrm{plan}}_t,o_t).
\end{equation}
This five-frame context captures task progress over the observed trajectory.

\textbf{Low-Level Action Generation.}
The low-level VLA policy adopts QwenPi~\citep{community2026starvla}, with its vision-language backbone initialized from Qwen3-VL-4B-Instruct.
Conditioned on the instruction, predicted subtask, and gripper state, it uses the initial observation, an observation from $\Delta=3$~s earlier, and the current observation to generate a flight action chunk:
\begin{equation}
\label{eq:agent-action}
\mathbf{A}_t=(a_t,\ldots,a_{t+H-1})
\sim\pi_\phi (\,\cdot\mid\ell,z_t,g_t,o_0,o_{t-\Delta},o_t),
\end{equation}
where $H$ is the chunk length and each flight action specifies a 6-DoF command.
The policy also determines when to invoke downstream manipulation, ending the current flight segment and transferring control for grasping or placing.

\textbf{Training.}
The planner and policy are trained independently.
The planner is fine-tuned on annotated subtasks using a language-modeling loss.
The policy learns action chunks through conditional flow matching.
Let $c_t=(\ell,z_t,g_t,o_0,o_{t-\Delta},o_t)$ denote the policy inputs.
Given a demonstrated action chunk $\mathbf{A}_t$, noise $\epsilon\sim\mathcal{N}(0,I)$, and flow time $\tau$, we construct
$\mathbf{A}_t^\tau=(1-\tau)\mathbf{A}_t+\tau\epsilon$
and minimize
\begin{equation}
\label{eq:flow-matching}
\mathcal{L}_{\mathrm{FM}}
=
\mathbb{E}_{\mathbf{A}_t,c_t,\epsilon,\tau}
\left[
\left\|
v_\phi(\mathbf{A}_t^\tau,\tau\mid c_t)
-(\epsilon-\mathbf{A}_t)
\right\|_2^2
\right],
\end{equation}
where $v_\phi$ predicts the conditional velocity field.
The policy additionally receives supervision for manipulation invocation.
It uses ground-truth subtasks during training and planner predictions at inference.
Each component uses the same visual sampling strategy during training and inference.
\subsection{More Qualitative Results}
\label{appendix:qualitative}
\begin{figure}[H]
  \centering
  \includegraphics[width=\linewidth]{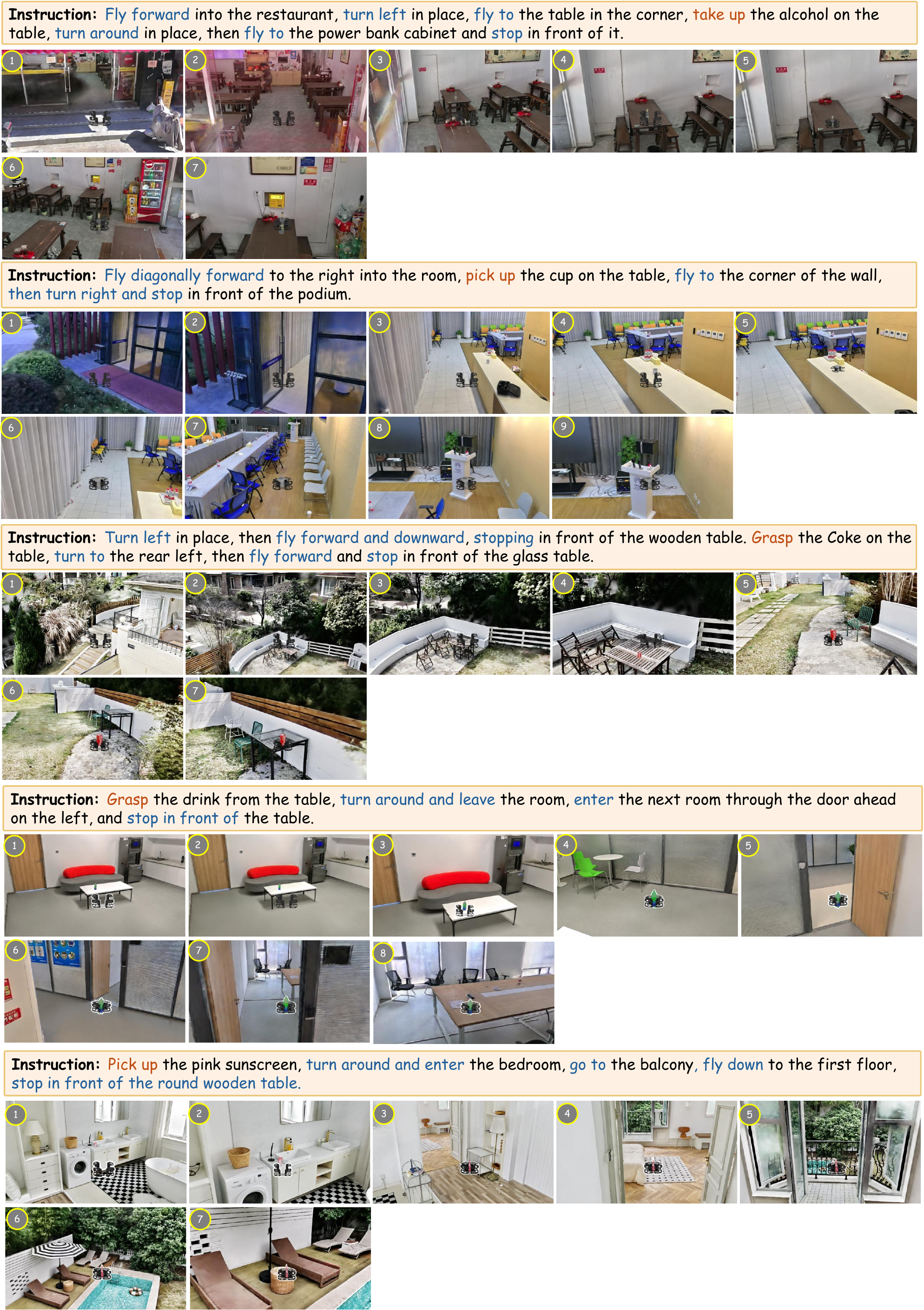}
  \vspace{-20pt}
  \caption{\textbf{Qualitative results of \textsc{\projectname-Agent}}. }
  \label{appendix:good-qualitative-results}
\end{figure}
\subsection{Failure Case}
\begin{figure}[H]
  \centering
  \includegraphics[width=\linewidth]{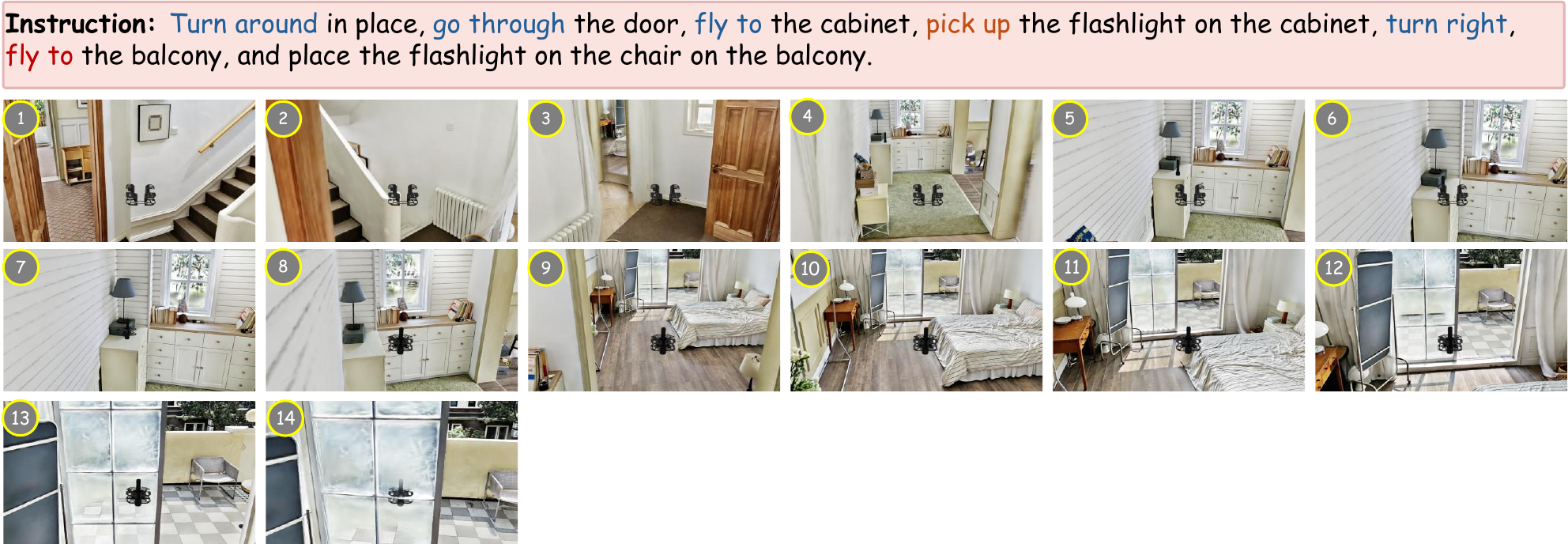}
  \vspace{-20pt}
  \caption{\textbf{Failure case.} Sometimes, UAV collides with obstacles when traversing passages across indoor--outdoor boundaries, indicating that navigation in transition regions remains challenging.}
  \label{appendix:embodiment-qualitative-results}
\end{figure}

\newcommand{\NTrainScenes}{18}
\newcommand{\NUnseenScenes}{5}

\section{Supplementary Information of Benchmark Splits}
\label{app:benchmark_split}

\subsection{Scene-Level Split}
\label{app:scene_split}

\projectname~contains 23 independently reconstructed scenes and 15,263
trajectories in total. We construct the benchmark split at the
scene level, rather than randomly splitting individual trajectories.
Let $\mathcal{S}$ denote the complete set of reconstructed scenes. As shown in Table~\ref{tab:scene_split}, we divide
it into a training-scene set $\mathcal{S}_{\mathrm{train}}$ and an
unseen-scene set $\mathcal{S}_{\mathrm{unseen}}$:
\begin{equation}
    \mathcal{S}
    =
    \mathcal{S}_{\mathrm{train}}
    \cup
    \mathcal{S}_{\mathrm{unseen}},
    \qquad
    \mathcal{S}_{\mathrm{train}}
    \cap
    \mathcal{S}_{\mathrm{unseen}}
    =
    \varnothing,
\end{equation}
where
$|\mathcal{S}_{\mathrm{train}}|=\NTrainScenes$ and
$|\mathcal{S}_{\mathrm{unseen}}|=\NUnseenScenes$, with
$\NTrainScenes+\NUnseenScenes=23$.
The split is fixed and shared by all evaluated methods.

Scene assignment is atomic: all trajectories associated with the
same reconstructed scene are assigned to the same scene partition,
regardless of whether they belong to the \textit{Indoor}, \textit{Outdoor}, or
\textit{Indoor$\leftrightarrow$Outdoor} task suite. In particular, trajectories
from different regions or task suites of the same reconstructed scene are
never separated between the training and unseen partitions.

The partition is therefore strictly isolated at the scene level. All
trajectories generated from scenes in
$\mathcal{S}_{\mathrm{unseen}}$, together with their RGB observations,
6-DoF UAV states, waypoint annotations, language instructions, and subtask
annotations, are excluded from model training. Consequently, no trajectory
or observation from an unseen scene is exposed to an agent during training.

\begingroup
\setlength{\intextsep}{6pt}
\begin{table}[!t]
  \setlength{\abovecaptionskip}{0cm}
  \setlength{\belowcaptionskip}{6pt}
  \centering
  \caption{\textbf{Scene-level partition used in \projectname~(\S\ref{app:scene_split}).}}
  \label{tab:scene_split}
  \resizebox{0.8\linewidth}{!}{%
    \setlength\tabcolsep{8pt}
    \renewcommand\arraystretch{1.02}
    \begin{tabular}{l||ccc}
      \hline\thickhline

      \rowcolor{mygray}
      \multicolumn{1}{c||}{\textbf{Split}}
      & \textbf{\# Scenes}
      & \textbf{Scene overlap with training}
      & \textbf{Used for training} \\
      \hline\hline

      Training
      & \NTrainScenes
      & --
      & Yes \\

      Seen
      & \NTrainScenes
      & Same scenes
      & No \\

      Unseen
      & \NUnseenScenes
      & None
      & No \\

      \hline
    \end{tabular}}
\end{table}
\endgroup

\subsection{Seen- and Unseen-Scene Evaluation}
\label{app:seen_unseen}

We distinguish between seen-scene and unseen-scene
evaluation according to whether the underlying reconstructed environment
is represented in the training set.

\textbf{Seen-Scene Evaluation.}
The seen split consists of held-out evaluation trajectories collected from
scenes in $\mathcal{S}_{\mathrm{train}}$. Although the underlying scenes
are represented during training, the evaluation trajectories themselves
are excluded from optimization. Denoting the training and seen evaluation
trajectory sets by $\mathcal{D}_{\mathrm{train}}$ and
$\mathcal{D}_{\mathrm{seen}}$, respectively, we enforce
\begin{equation}
    \mathcal{D}_{\mathrm{train}}
    \cap
    \mathcal{D}_{\mathrm{seen}}
    =
    \varnothing.
\end{equation}
Therefore, seen-scene evaluation measures generalization to new episodes,
routes, and task configurations within environments observed during
training, rather than performance on trajectories used for optimization.

\textbf{Unseen-Scene Evaluation.}
The unseen split consists exclusively of trajectories collected from
$\mathcal{S}_{\mathrm{unseen}}$. Let
$\mathcal{D}_{\mathrm{unseen}}$ denote the corresponding evaluation
trajectory set. By construction,
\begin{equation}
    \operatorname{scene}(\mathcal{D}_{\mathrm{train}})
    \cap
    \operatorname{scene}(\mathcal{D}_{\mathrm{unseen}})
    =
    \varnothing.
\end{equation}
Thus, unseen-scene evaluation introduces both trajectory-level and
scene-level separation from the training data. Agents are evaluated in
reconstructed environments whose visual observations, scene geometry, and
spatial layouts are not encountered during training. This setting measures
generalization to novel environments rather than to new trajectories
within familiar scenes.

\section{Implementation Details}
\label{app:implementation}
For reproducibility, we provide the detailed training configuration used for all VLA models. Unless otherwise specified, all VLA models follow the same training protocol, as summarized in Table~\ref{tab:vla-training-config}.

\begingroup
\setlength{\intextsep}{6pt}
\begin{table}[H]
  \setlength{\abovecaptionskip}{0cm}
  \setlength{\belowcaptionskip}{6pt}
  \centering
  \caption{\textbf{Training configuration for VLA models.}}
  \label{tab:vla-training-config}
  \resizebox{0.55\linewidth}{!}{%
    \setlength\tabcolsep{8pt}
    \renewcommand\arraystretch{1.02}
    \begin{tabular}{l||c}
      \hline\thickhline

      \rowcolor{mygray}
      \multicolumn{1}{c||}{\textbf{Training Configuration}}
      & \textbf{Setting} \\
      \hline\hline

      Batch size
      & 32 \\

      Learning rate
      & $1\times10^{-5}$ \\

      Learning rate schedule
      & Warmup + cosine decay \\

      Optimizer
      & AdamW \\

      AdamW $\beta$
      & $(0.9, 0.95)$ \\

      Weight decay
      & $1\times10^{-8}$ \\

      Max. training steps
      & 100,000 \\

      Action horizon
      & 36 \\

      Precision
      & bf16 \\

      Training hardware
      & $4\times$ NVIDIA H100 GPUs \\

      \hline
    \end{tabular}}
\end{table}
\endgroup

We apply this training protocol consistently to all VLA-based models in our experiments.

\section{Discussion}
\label{Discussion}
Aerial mobile manipulation naturally couples two stages operating at different spatial and control scales: long-range mobility and local physical interaction. While mobility requires UAV to follow instructions, traverse potentially heterogeneous environments, and approach a target, manipulation is typically executed after the platform reaches a suitable pose. The key challenge is the handoff from mobility to manipulation. \projectname~makes this handoff explicit through interaction-oriented mobility and the Interaction Readiness Rate (IRR), which evaluates whether the terminal navigation state satisfies the spatial and viewpoint conditions required for downstream interaction. Once such a state is reached, an interaction-invocation signal can trigger a task-specific manipulation policy.

This formulation provides a modular pathway for extending aerial VLA agents toward mobile manipulation without requiring a single policy to jointly learn long-horizon flight and local interaction. Existing aerial VLA agents can retain their role in instruction following and target approach, while downstream policies handle grasping or other local skills. Our manipulation experiments provide initial evidence for this interface: under the same grasping policy, interaction-ready states can support successful manipulation, whereas unsuitable arrival states may lead to failure. More broadly, this modular view enables mobility and manipulation to be developed independently while remaining connected within a unified embodied execution pipeline.
\end{document}

%% file: math_commands.tex
\usepackage{amsmath,amsfonts,bm}

\def\eqref#1{equation~\ref{#1}}

\def\1{\bm{1}}

\DeclareMathAlphabet{\mathsfit}{\encodingdefault}{\sfdefault}{m}{sl}
\SetMathAlphabet{\mathsfit}{bold}{\encodingdefault}{\sfdefault}{bx}{n}

